\documentclass[11pt,a4paper]{article}

\usepackage[hyperref]{naaclhlt2018}
\usepackage{times}
\usepackage{latexsym}

\usepackage{url}

\usepackage{epsfig}
\usepackage{graphicx}
\usepackage{amsmath}
\usepackage{amssymb}
\usepackage{subfig}

\aclfinalcopy 
\def\aclpaperid{3050} 

\title{Neural Machine Translation for Low Resource Languages using Bilingual Lexicon Induced from Comparable Corpora}

\author{Sree Harsha Ramesh \and Krishna Prasad Sankaranarayanan \\
College of Information and Computer Sciences \\
University of Massachusetts Amherst \\
{\tt \{shramesh, ksankaranara\}@cs.umass.edu}}

\date{}

\begin{document}
\maketitle

\begin{abstract}
   Resources for the non-English languages are scarce and this paper addresses this problem in the context of machine translation, by automatically extracting parallel sentence pairs from the multilingual articles available on the Internet. In this paper, we have used an end-to-end Siamese bidirectional recurrent neural network to generate parallel sentences from comparable multilingual articles in Wikipedia.  Subsequently, we have showed that using the harvested dataset improved BLEU scores on both NMT and phrase-based SMT systems for the low-resource language pairs: English--Hindi and English--Tamil, when compared to training exclusively on the limited bilingual corpora collected for these language pairs.
\end{abstract}

\section{Introduction}

\begin{figure*}
  \includegraphics[width=16cm,height=10cm]{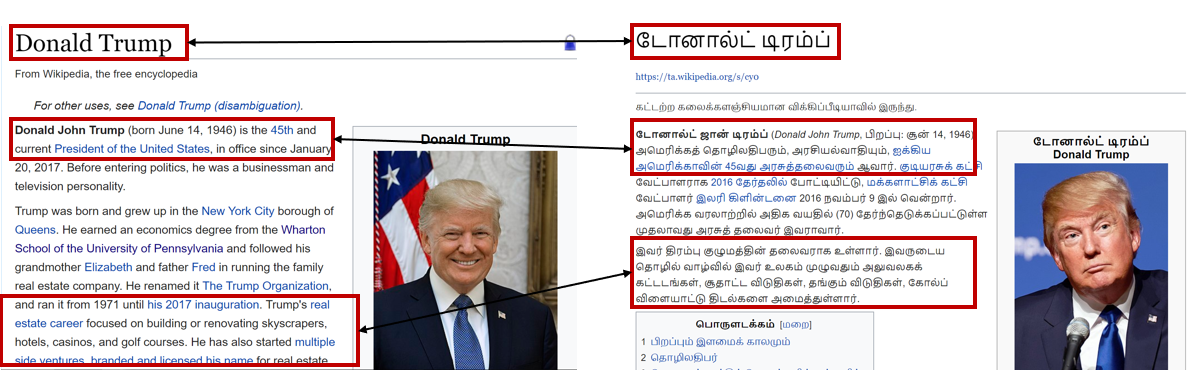}
  \caption{A side-by-side comparison of nearly parallel sentences from bilingual Wikipedia articles about Donald Trump in English and Tamil.}
  \label{fig:trump}
\end{figure*}

Both neural and statistical machine translation approaches are highly reliant on the availability of large amounts of data and are known to perform poorly in low resource settings. Recent crowd-sourcing efforts and workshops on machine translation have resulted in small amounts of parallel texts for building viable machine translation systems for low-resource pairs \cite{post2012constructing}. But, they have been shown to suffer from low accuracy (incorrect translation) and low coverage (high out-of-vocabulary rates), due to insufficient training data. In this project, we try to address the high OOV rates in low-resource machine translation systems by leveraging the increasing amount of multilingual content available on the Internet for enriching the bilingual lexicon. 

Comparable corpora such as Wikipedia, are collections of topic-aligned but non-sentence-aligned multilingual documents which are rich resources for extracting parallel sentences from. For example, Figure \ref{fig:trump} shows that there are equivalent sentences on the page about Donald Trump in Tamil and English, and the phrase alignment for an example sentence is shown in Table \ref{table:word-table}.
\begin{table}[]
\centering
\begin{tabular}{|l|l|l|}
\hline
\textbf{\begin{tabular}[c]{@{}l@{}}Language\\  (\textit{ISO 639-1})\end{tabular}} & \textbf{\begin{tabular}[c]{@{}l@{}}\# Bilingual\\  Wiki articles\end{tabular}} & \textbf{\begin{tabular}[c]{@{}l@{}}\# Curated \\ en--xx \\sent. pairs\\ \end{tabular}} \\ \hline
Urdu (\textit{ur})              & 124,078                                                                         & 35,916                                                           \\ \hline
Hindi (\textit{hi})             & 121,234                                                                         & 1,495,854                                                                              \\ \hline
Tamil (\textit{ta})            & 113,197                                                                         & 169,871                                                                                         \\ \hline
Telugu (\textit{te})           & 67,508                                                                          & 46,264                                                                                          \\ \hline
Bengali (\textit{bn})          & 52,518                                                                          & 23,610                                                                                          \\ \hline
Malayalam (\textit{ml})         & 52,224                                                                          & 33,248                                                                                          \\ \hline
\end{tabular}
\bigskip
\caption{Number of bilingual articles in Wikipedia against the number of parallel sentences in the largest xx--en corpora available.}
\label{table:wiki-count}
\end{table}

Table \ref{table:wiki-count} shows that there are at least tens of thousands of bilingual articles on Wikipedia which could potentially have at least as many parallel sentences that could be mined to address the scarcity of parallel sentences as indicated in column 2 which shows the number of sentence-pairs in the largest available bilingual corpora for xx-en\footnote{en--ta : http://ufal.mff.cuni.cz/\textasciitilde{}ramasamy/parallel/html/ \\en--hi:  http://www.cfilt.iitb.ac.in/iitb\_parallel/ \\ en--others:https://github.com/joshua-decoder/indian-parallel-corpora}. As shown by \citeauthor{irvine2013combining} (\citeyear{irvine2013combining}), the illustrated data sparsity can be addressed by extending the scarce parallel sentence-pairs with those automatically extracted from Wikipedia and thereby improving the performance of statistical machine translation systems. 

In this paper, we will propose a neural approach to parallel sentence extraction and compare the BLEU scores of machine translation systems with and without the use of the extracted sentence pairs to justify the effectiveness of this method. Compared to previous approaches which require specialized meta-data from document structure or significant amount of hand-engineered features, the neural model for extracting parallel sentences is learned end-to-end using only a small bootstrap set of parallel sentence pairs.
\begin{table}
  \includegraphics[width=7.5cm]{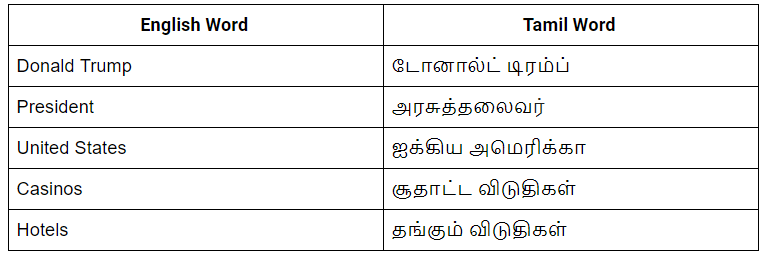}
  \caption{Phrase-aligned en--ta pairs from Fig \ref{fig:trump}}
  \label{table:word-table}
\end{table}
\section{Related Work}
A lot of work has been done on the problem of automatic sentence alignment from comparable corpora, but a majority of them \citep{abdui2009use, irvine2013combining,yasuda2008method} use a pre-existing translation system as a precursor to ranking the candidate sentence pairs, which the low resource language pairs are not at the luxury of having; or use statistical machine learning approaches, where a Maximum Entropy classifier is used that relies on surface level features such as word overlap in order to obtain parallel sentence pairs \citep{munteanu2005improving}. However, the deep neural network model used in our paper is probably the first of its kind, which does not need any feature engineering and also does not need a pre-existing translation system. 


\citeauthor{munteanu2005improving} (\citeyear{munteanu2005improving}) proposed a parallel sentence extraction system which used comparable corpora from newspaper articles to extract the parallel sentence pairs. In this procedure, a maximum entropy classifier is designed for all sentence pairs possible from the Cartesian product of a pair of documents and passed through a sentence-length ratio filter in order to obtain candidate sentence pairs. SMT systems were trained on the extracted sentence pairs using the additional features from the comparable corpora like distortion and position of current and previously aligned sentences. This resulted in a state of the art approach with respect to the translation performance of low resource languages. 

Similar to our proposed approach, \citeauthor{barron2015factory} (\citeyear{barron2015factory}) showed how using parallel documents from Wikipedia for domain specific alignment would  improve translation quality of SMT systems on in-domain data. In this method, similarity between all pairs of cross-language sentences with different text similarity measures are estimated. The issue of domain definition is overcome by the use of IR techniques which use the characteristic vocabulary of the domain to query a Lucene search engine over the entire corpus. The candidate sentences are defined based on word overlap and the decision whether a sentence pair is parallel or not using the maximum entropy classifier. The difference in the BLEU scores between out of domain and domain-specific translation is proved clearly using the word embeddings from characteristic vocabulary extracted using the extracted additional bitexts.

\citeauthor{abdui2009use} (\citeyear{abdui2009use}) extract parallel sentences without the use of a classifier. Target language candidate sentences are found using the translation of source side comparable corpora. Sentence tail removal is used to strip the tail parts of sentence pairs which differ only at the end. This, along with the use of parallel sentences enhanced the BLEU score and helped to determine if the translated source sentence and candidate target sentence are parallel by measuring the word and translation error rate. This method succeeds in  eliminating the need for domain specific text by using the target side as a source of candidate sentences. However, this approach is not feasible if there isn't a good source side translation system to begin with, like in our case.

Yet another approach which uses an existing translation system to extract parallel sentences from comparable documents was proposed by \citeauthor{yasuda2008method} (\citeyear{yasuda2008method}). They describe a framework for machine translation using multilingual Wikipedia articles. The parallel corpus is assembled iteratively, by using a statistical machine translation system trained on a preliminary sentence-aligned corpus, to score sentence-level en--jp BLEU scores. After filtering out the unaligned pairs based on the MT evaluation metric, the SMT is retrained on the filtered pairs. 
\section{Approach}
\begin{figure*}
  \includegraphics[width=17.5cm,height=10cm]{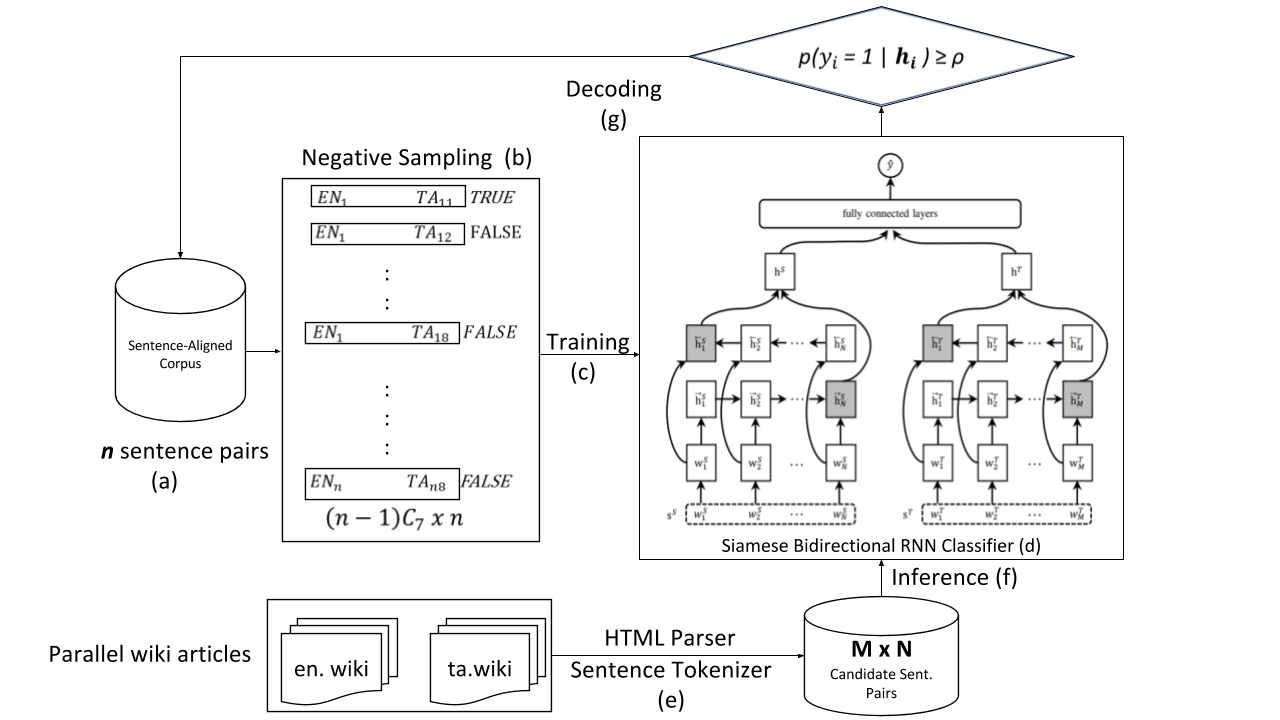}
  \caption{Architecture for the parallel sentence extraction system showing training and inference pipelines. EN - English, TA - Tamil}
  \label{fig:architecture}
\end{figure*}

In this section, we will describe the entire pipeline, depicted in Figure \ref{fig:architecture}, which is involved in training a parallel sentence extraction system, and also to infer and decode high-precision nearly-parallel sentence-pairs from bilingual article pages collected from Wikipedia. 

\subsection{Bootstrap Dataset}
The parallel sentence extraction system needs a sentence aligned corpus which has been curated. These sentences were used as the ground truth pairs when we trained the model to classify parallel sentence pair from non-parallel pairs.

\subsection{Negative Sampling} \label{neg-sampling}
The binary classifier described in the next section, assigns a translation probability score to a given sentence pair, after learning from examples of translations and negative examples of non-translation pairs. For, this we make a simplistic assumption that the parallel sentence pairs found in the bootstrap dataset are unique combinations, which fail being translations of each other, when we randomly pick a sentence from both the sets. Thus, there might be cases of false negatives due to the reliance on unsupervised random sampling  for generation of negative labels. 

Therefore at the beginning of every epoch, we randomly sample $m$ negative sentences of the target language for every source sentence. From a few experiments and also from the literature, we converged on $m=7$ to be performing the best, given our compute constraints.  

\subsection{Model}
Here, we describe the neural network architecture as shown in \citeauthor{gregoire2017deep} (\citeyear{gregoire2017deep}), where the network learns to estimate the probability that the sentences in a given sentence pair, are translations of each other, \(p(y_{i}=1|\boldsymbol{s}^{S}_{i},\boldsymbol{s}^{T}_{i})\), where $\boldsymbol{s}_{i}^{S}$ is the candidate source sentence in the given pair, and $\boldsymbol{s}_{i}^{T}$ is the candidate target sentence. 
\subsubsection{Training}
As illustrated in Figure \ref{fig:architecture} (d), the architecture uses a siamese network \cite{bromley1994signature}, consisting of a bidirectional RNN \cite{schuster1997bidirectional} sentence encoder with recurrent units such as long short-term memory units, or LSTMs \cite{hochreiter1997long} and gated recurrent units, or GRUs \cite{cho2014learning} learning a vector representation for the source and target sentences and the probability of any given pair of sentences being translations of each other. For seq2seq architectures, especially in translation, we have found the that the recommended recurrent unit is GRU, and all our experiments use this over LSTM. 

The forward RNN reads the variable-length sentence and updates its recurrent state from the first token until the last one to create a fixed-size continuous vector representation of the sentence. The backward RNN processes the sentence in reverse. In our experiments, we use the concatenation of the last recurrent state in both directions as a final representation \(\mathbf{h}^{S}_{i}\ = [\overrightarrow{\mathbf{h}}^{S}_{i,N}\ ; \overleftarrow{\mathbf{h}}^{S}_{i,1}]\)
\begin{gather}
  \mathbf{w}^{S}_{i,t} = \mathbf{E}^{S^\top} \mathbf{w}_{k} \\
  \overrightarrow{\mathbf{h}}^{S}_{i,t} = \phi(\overrightarrow{\mathbf{h}}^{S}_{i, t-1},
    \mathbf{w}^{S}_{i,t}) \\
  \overleftarrow{\mathbf{h}}^{S}_{i,t} = \phi(\overleftarrow{\mathbf{h}}^{S}_{i, t+1},
    \mathbf{w}^{S}_{i,t})
\end{gather}
where \(\phi\) is the gated recurrent unit (GRU). After both source and target sentences have been encoded, we capture their matching information by using their element-wise product and absolute element-wise difference. We estimate the probability that the sentences are translations of each other by feeding the matching vectors into fully connected layers:
\begin{gather}
  \mathbf{h}_{i}^{(1)} = \mathbf{h}^{S}_{i} \odot \mathbf{h}^{T}_{i} \\
  \mathbf{h}_{i}^{(2)} = |\mathbf{h}^{S}_{i} - \mathbf{h}^{T}_{i}| \\
  \mathbf{h}_{i} = tanh(\mathbf{W}^{(1)}\mathbf{h}_{i}^{(1)} + \mathbf{W}^{(2)}\mathbf{h}_{i}^{(2)} + \mathbf{b}) \\
  p(y_{i}=1|\mathbf{h}_{i}) = \sigma(\mathbf{W}^{(3)}\mathbf{h}_{i} + c)
\end{gather}
where \(\sigma\) is the sigmoid function, \(\mathbf{W}^{(1)}\), \(\mathbf{W}^{(2)}\), \(\mathbf{W}^{(3)}\), \(\mathbf{b}\) and \(c\) are model parameters. The model is trained by minimizing the cross entropy of our labeled sentence pairs:
\begin{equation}
  \begin{split}
   \mathcal{L}
   = &-\sum^{n(1+m)}_{i=1} y_{i} \log \sigma(\mathbf{W}^{(3)}\mathbf{h}_{i} + c) \\
     &-(1-y_{i}) \log (1-\sigma(\mathbf{W}^{(3)}\mathbf{h}_{i} + c))
  \end{split}
\end{equation}
where $n$ is the number of source sentences and $m$ is the number of candidate target sentences being considered. 
\begin{table*}
 \centering 
 \includegraphics[width=\textwidth]{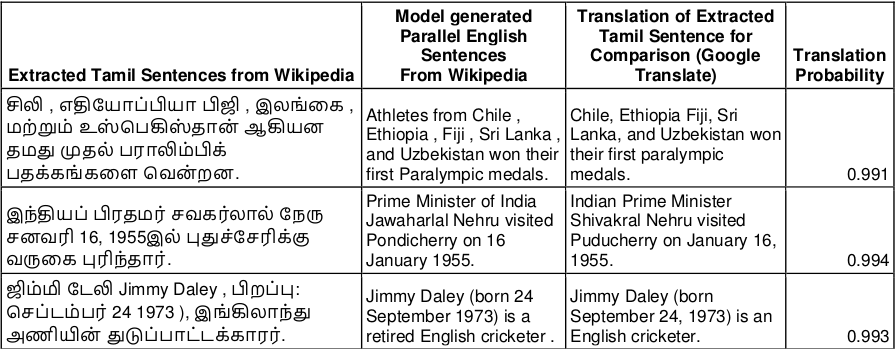}
 \caption{A sample of parallel sentences extracted from Wiki en--ta articles. The translation of the extracted Tamil sentence in English is also provided. Translation probability corresponds to our model's score of how likely the sentences are translations of each other, as calculated in Equation 8.}
\label{table:tamil-wiki-sents}
\end{table*}
\subsubsection{Inference}
For prediction, a sentence pair is classified as parallel if the probability score is greater than or equal to a decision threshold \(\rho\) that we need to fix. We found that to get high precision sentence pairs, we had to use $\rho=0.99$, and if we were able to sacrifice some precision for recall, a lower $\rho=0.80$ of 0.80 would work in the favor of reducing OOV rates.
\begin{equation}
  \hat{y}_{i} =
    \begin{cases}
      1 & \text{if}\ p(y_{i}=1|\mathbf{h}_{i}) \geq \rho \\
      0 & \text{otherwise}
    \end{cases}
\end{equation}

\begin{figure*}[t]
    \centering
    \subfloat[Without greedy decoding]  {\includegraphics[width=7cm]{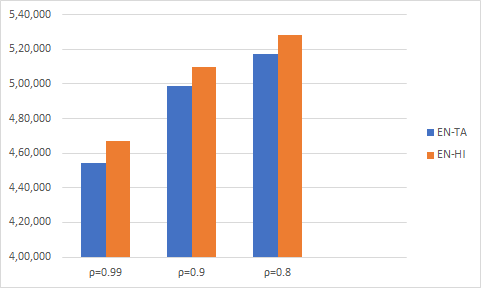}
    \label{fig:greedy-false}
    }\qquad
    \subfloat[With greedy decoding]{\includegraphics[width=7cm]{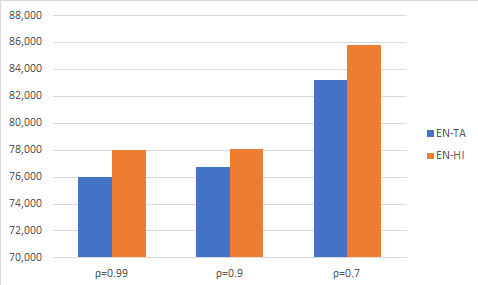} 
    \label{fig:greedy-true}
    }
    \caption{Number of parallel sentences extracted from 10,000 parallel Wikipedia article pairs using different thresholds and decoding methods}%
    \label{fig:example}%
\end{figure*}

\section{Experiments}

\subsection{Dataset}
We experimented with two language pairs: English -- Hindi (en--hi) and English -- Tamil (en--ta). The parallel sentence extraction systems for both en--ta and en--hi were trained using the architecture described in \ref{neg-sampling} on the following bootstrap set of parallel corpora: 
\begin{itemize}
\item An English-Tamil parallel corpus \cite{ramasamy2014entam} containing a total of $169,871$ sentence pairs, composed of $3,984,038$ English Tokens and $2,776,397$ Tamil Tokens.
\item An English-Hindi parallel corpus \cite{kunchukuttan2017iit} containing a total of $1,492,827$ sentence pairs, from which a set of $200,000$ sentence pairs were picked randomly.
\end{itemize}
Subsequently, we extracted parallel sentences using the trained model, and  parallel articles collected from Wikipedia\footnote{Tamil: dumps.wikimedia.org/tawiki/latest/ \\ Hindi: dumps.wikimedia.org/hiwiki/latest/}. There were $67,449$ bilingual English-Tamil and $58,802$ English-Hindi titles on the Wikimedia dumps collected in December 2017. 

\subsection{Evaluation Metrics}
For the evaluation of the performance of our sentence extraction models, we looked at a few sentences manually, and have done a qualitative analysis, as there was no gold standard evaluation set for sentences extracted from Wikipedia. In Table \ref{table:tamil-wiki-sents}, we can see the qualitative accuracy for some parallel sentences extracted from Tamil. The sentences extracted from Tamil, have been translated to English using Google Translate, so as to facilitate a comparison with the sentences extracted from English. 

For the statistical machine translation and neural machine translation evaluation we use the BLEU score \cite{papineni2002bleu} as an evaluation metric, computed using the \textit{multi-bleu} script from Moses \cite{koehn2007moses}.

\subsection{Sentence Alignment}
Figures \ref{fig:greedy-false} shows the number of high precision sentences that were extracted at $\rho=0.99$ without greedy decoding. Greedy decoding could be thought of as sampling without replacement, where a sentence that's already been extracted on one side of the extraction system, is precluded from being considered again. Hence, the number of sentences without greedy decoding, are of an order of magnitude higher than with decoding, as can be seen in Figure \ref{fig:greedy-true}.

\begin{table*}
\centering
\begin{tabular}{cccc}
\hline
\textbf{Training Data}  & \textbf{Model} & \textbf{BLEU} & \textbf{\#Sents} \\ \hline
IIT Bombay en--hi        & SMT            & 2.96          & 200,000              \\ 
+ Wiki Extracted ρ=0.99 & SMT            & 3.57(+0.61)   & +77,988               \\ \hline
IIT Bombay en--hi        & NMT            & 3.46          & 200,000              \\ 
+ Wiki Extracted ρ=0.99 & NMT            & 3.97(+0.51)   & +77,988               \\ \hline
Ramasamy et.al  en--ta   & SMT            & 4.02          & 169,871              \\ 
+ Wiki Extracted ρ=0.99 & SMT            & 4.57(+0.55)   & +75,970               \\ \hline
Ramasamy et.al  en--ta   & NMT            & 4.53          & 169,871              \\ 
+ Wiki Extracted ρ=0.99 & NMT            & 5.03(+0.50)   & +75,970               \\ \hline
\end{tabular}
\caption{BLEU score results for en--hi and en--ta}
\label{table:bleu-hi}
\end{table*}

\subsection{Machine Translation}
We evaluated the quality of the extracted parallel sentence pairs, by performing machine translation experiments on the augmented parallel corpus.
\subsubsection{SMT}
As the dataset for training the machine translation systems, we used high precision sentences extracted with greedy decoding, by ranking the sentence-pairs on their translation probabilities. Phrase-Based SMT systems were trained using Moses \cite{koehn2007moses}. We used the \textit{grow-diag-final-and} heuristic for extracting phrases, lexicalised reordering and Batch MIRA \cite{cherry2012batch} for tuning (the default parameters on Moses). We trained 5-gram language models with Kneser-Ney smoothing using KenLM \cite{heafield2013scalable}. With these parameters, we trained SMT systems for en--ta and en--hi language pairs, with and without the use of extracted parallel sentence pairs.

\subsubsection{NMT}
For training neural machine translation models, we used the TensorFlow \cite{abadi2016tensorflow} implementation of OpenNMT \cite{2017opennmt} with attention-based transformer architecture \cite{vaswani2017attention}. The BLEU scores for the NMT models were higher than for SMT models, for both en--ta and en--hi pairs, as can be seen in Table \ref{table:bleu-hi}.

\section{Conclusion}

In this paper, we evaluated the benefits of using a neural network procedure to extract parallel sentences. Unlike traditional translation systems which make use of multi-step classification procedures, this method requires just a parallel corpus to extract parallel sentence pairs using a Siamese BiRNN encoder using GRU as the activation function.
 
This method is extremely beneficial for translating language pairs with very little parallel corpora. These parallel sentences facilitate significant improvement in machine translation quality when compared to a generic system as has been shown in our results. 

The experiments are shown for English-Tamil and English-Hindi language pairs. Our model achieved a marked percentage increase in the BLEU score for both en--ta and en--hi language pairs. We demonstrated a percentage increase in BLEU scores of 11.03\% and 14.7\% for en--ta and en--hi pairs respectively, due to the use of parallel-sentence pairs extracted from comparable corpora using the neural architecture.

As a follow-up to this work, we would be comparing our framework against other sentence alignment methods described in \cite{resnik2003web}, \cite{Ayan:2006:GBA:1220175.1220177}, \cite{rosti2007combining} and  \cite{smith2010extracting}. It has also been interesting to note that the 2018 edition of the Workshop on Machine Translation (WMT) has released a new shared task called Parallel Corpus Filtering \footnote{http://statmt.org/wmt18/parallel-corpus-filtering.html} where participants develop methods to filter a given noisy parallel corpus (crawled from the web), to a smaller size of high quality sentence pairs. This would be the perfect avenue to test the efficacy of our neural network based approach of extracting parallel sentences from unaligned corpora.  

\bibliography{naaclhlt2018}
\bibliographystyle{acl_natbib}

\end{document}